\documentclass[sigconf]{acmart}

\usepackage{multirow}
\usepackage{float}
\usepackage{fancyvrb}
\usepackage{amsmath}
\usepackage{amsfonts}
\usepackage{array}
\usepackage{ragged2e}
\usepackage{tabu}
\usepackage{algorithmicx}
\usepackage{algorithm}
\usepackage{algpseudocode}

\AtBeginDocument{%
  \providecommand\BibTeX{{%
    \normalfont B\kern-0.5em{\scshape i\kern-0.25em b}\kern-0.8em\TeX}}}

\setcopyright{acmcopyright}
\copyrightyear{2022}
\acmYear{2022}
\acmDOI{}

\acmConference[]{}{}{}
\acmPrice{0}
\acmISBN{}

\acmSubmissionID{785}

\begin{document}

\title[MHMS: Multimodal Hierarchical Multimedia Summarization]{MHMS: Multimodal Hierarchical Multimedia Summarization}

\author{Jielin Qiu$^{1}$, ~Jiacheng Zhu$^{1}$, ~Mengdi Xu$^{1}$, ~Franck Dernoncourt$^{2}$, \\
~Zhaowen Wang$^{2}$,  ~Trung Bui$^{2}$, ~Bo Li$^{3}$, ~Ding Zhao$^{1}$, ~Hailin Jin$^{2}$}
\affiliation{%
  \institution{$^{1}$Carnegie Mellon University, 
  ~$^{2}$Adobe Research, 
  ~$^{3}$University of Illinois Urbana-Champaign \\ 
  $\left\{\text{jielinq,jzhu4,mengdixu}\right\}$@andrew.cmu.edu, $\left\{\text{dernonco,zhawang,bui,hljin}\right\}$@adobe.com,
  lbo@illinois.edu
  }
  \country{}
}

\renewcommand{\shortauthors}{}

\begin{abstract}
  Multimedia summarization with multimodal output can play an essential role in real-world applications, i.e., automatically generating cover images and titles for news articles or providing introductions to online videos. In this work, we propose a multimodal hierarchical multimedia summarization (MHMS) framework by interacting visual and language domains to generate both video and textual summaries. Our MHMS method contains video and textual segmentation and summarization module, respectively. It formulates a cross-domain alignment objective with optimal transport distance which leverages cross-domain interaction to generate the representative keyframe and textual summary. We evaluated MHMS on three recent multimodal datasets and demonstrated the effectiveness of our method in producing high-quality multimodal summaries.
\end{abstract}

\ccsdesc[500]{Information systems~Summarization}
\ccsdesc[500]{Information systems~Multimedia and multimodal retrieval}

\keywords{Multimodal summarization, video temporal segmentation, video summarization, textual segmentation, textual summarization, cross-domain alignment, optimal transport}


\maketitle

\section{Introduction}
New multimedia contents in the form of short videos and corresponding text articles have become a major trend on influential digital media including CNN, BBC, Daily Mail, social media, etc \cite{Mingzhe2020VMSMOLT}.
This popular media type has shown to be successful in drawing user attention and delivering key information in a short time.
The summarization of multimedia data is also becoming increasingly important in real-world applications such as automatically generating cover images and titles for news articles and providing introductions to online videos. Summarization of multimedia aims to extract the most important information from a variety of conceptually related media sources, so that a short, concise and informative version of the original contents is produced.

Multimedia summarization can be divided into three categories based on domains, including video summarization, textual summarization, and multimodal summarization.
Video summarization aims to generate a short synopsis to summarize the video content by selecting the most informative and essential information, where the summary is usually composed of a set of representative keyframes. Textual summarization targets at producing a concise and fluent summary while preserving critical information and overall meaning for the articles or documents. With the increasing interests in multimodal learning, multimodal summarization is becoming more popular, providing users with both visual and textual representative information, effectively improving the user experience.

Most existing video summarization methods used visual features from videos, thereby leaving out abundant information. 
For instance, \cite{Gygli2014CreatingSF,Jadon2020UnsupervisedVS} generated video summaries by selecting keyframes using SumMe and TVSum datasets. However, textual summarization of videos is less explored. Pure textual summarization only takes textual metadata, i.e., documents, articles, tweets, etc, as input, and generates textual only summaries. With the development in multimodal machine learning, incorporating additional modality into the learning process has drawn increasing attention \cite{Duan2022MultimodalAU}.
Some recent work proposed the query-based video summarization task, where additional text information for each video is given, i.e., category, search query, title, or description \cite{Haopeng2022VideoSB}. The method still focused on generating pure visual summaries. \cite{Sah2017SemanticTS} proposed to generate both visual and textual summaries of long videos using recurrent networks. However, the previous works tried to learn the whole representation for the entire video and articles, which leads to a constrain, since different parts of the video have different meanings, same for the articles.
We believe a segment-based multimodal summarization approach can provide more accurate summaries for the given multimedia source, which could also improve user satisfaction with the informativeness of summaries \cite{Zhu2018MSMOMS}.

In this work, we focus on the multimodal multimedia summarization task with a multimodal output, where we explore segment-based cross-domain representations through multimodal interactions to generate both visual and textual summaries. We divide the whole pipeline into sub-modules to handle the segmentation and summarization task within visual and textual domains, respectively. Then we use optimal transport as the bridge to align the representations from different modalities to generate the final multimodal summary.
Our contributions include:
\begin{itemize}
    \item We propose MHMS, a multimodal hierarchical multimedia summarization framework to generate both video and textual summaries for multimedia sources.
    \item Our method learns a joint representation and aligns cross-domain features to exploit the interaction between hybrid media types via optimal transport.
    \item Our experimental results on three public datasets demonstrate the effectiveness of our method compared with existing approaches, which could be adopted in many real-world applications.
\end{itemize}

\section{Related Work}
\paragraph{\textbf{Video Summarization}} 
Video Summarization aims at generating a short synopsis that summarizes the video content by selecting the most informative and vital parts. The methods lie in two directions: unimodal and multimodal approaches. Unimodal approaches only use the visual modality of the videos to learn summarization in a supervised manner, while multimodal methods exploit the available textual metadata and learn semantic or category-driven summarization in an unsupervised way. The summary usually contains a set of representative video keyframes or video key-fragments that have been stitched in chronological order to form a shorter video. The former type of video summary is called video storyboard, and the latter one is called video skim \cite{Apostolidis2021VideoSU}. Traditional video summarization methods only use visual information, trying to extract important frames to represent the video content. Some category-driven or supervised training approaches were proposed to generate video summaries with video-level labels \cite{song2015tvsum,zhou2018deep,xiao2020convolutional,Zhou2018VideoSB}.

\paragraph{\textbf{Textual Summarization}}
There are two language summarization methods: abstractive summarization and extractive summarization. Abstractive methods select words based on semantic understanding, and even the words may not appear in the source \cite{tan2017abstractive,see2017get}. Extractive methods attempt to summarize language by selecting a subset of words that retain the most critical points, which weights the essential part of sentences to form the summary \cite{narayan2018ranking, wu2018learning}. 
Recently, the fine-tuning approaches have improved the quality of generated summaries based on pre-trained language models in a wide range of tasks \citep{liu-lapata-2019-text, zhang-etal-2019-hibert}.

\paragraph{\textbf{Multimodal Summarization}}
Multimodal summarization is to exploit multiple modalities for summarization, i.e., audio signals, video captions, Automatic Speech Recognition (ASR) transcripts, video titles, or other contextual data. Some work tried to learn the relevance or mapping in the latent space between different modalities with trained models \cite{Otani2016VideoSU,Yuan2019VideoSB,Wei2018VideoSV,Fu2020MultimodalSF}.
In addition to only generating visual summaries, some work learned to generate textual summaries by taking audio, transcripts, or documents as input along with videos or images \cite{Li2017MultimodalSF,Atri2021SeeHR,Zhu2018MSMOMS}, using seq2seq model \cite{Sutskever2014SequenceTS} or attention mechanism \cite{Bahdanau2015NeuralMT}. The methods above explored using multiple modalities' information to generate single modality output, either textual or visual summary.

\paragraph{\textbf{Video Temporal Segmentation}} 
Video temporal segmentation aims at generating small video segments based on the content or topics of the video,  which is a fundamental step in content-based video analysis and plays a crucial role in video analysis.
Previous work mostly formed a classification problem to detect the segment boundaries in the supervised manner \cite{Sidiropoulos2011TemporalVS,Zhou2013HierarchicalAC,Poleg2014TemporalSO,Sokeh2018SuperframesAT,Aakur2019APP}. Recently, unsupervised methods have also been explored \cite{Gygli2014CreatingSF,Song2015TVSumSW}. 
Temporal segmentation of actions in videos has been widely explored in previous works \cite{Wang2019TemporalSN,Zhao2017TemporalAD,Lea2017TemporalCN,Kuehne2020AHR,Sarfraz2021TemporallyWeightedHC,Wang2020BoundaryAwareCN}. Video shot boundary detection and scene detection tasks are also relevant and has been explored in many previous studies \cite{Hassanien2017LargescaleFA,Hato2019FastAF,Rao2020ALA,Chen2021ShotCS,Zhang2021BetterLS}, which aim at finding the visual change or scene boundaries.

\paragraph{\textbf{Textual Segmentation}} 
Textual segmentation aim at dividing the text into coherent, contiguous, and semantically meaningful segments \cite{Nicholls2021ANM}. These segments can be composed of words, sentences, or topics, where the types of text include blogs, articles, news, video transcript, etc.
Previous work focused on heuristics-based methods \cite{Koshorek2018TextSA,Choi2000AdvancesID}, LDA-based modeling algorithms \cite{Blei2003LatentDA,Chen2009GlobalMO}, or Bayesian methods \cite{Chen2009GlobalMO,Riedl2012TopicTilingAT}. Recent developments in natural language processing developed large models to learn huge amount of data in the supervised manner \cite{Mikolov2013LinguisticRI,Pennington2014GloVeGV,Li2018SegBotAG,Wang2018TowardFA}. Besides, unsupervised or weakly-supervised methods has also drawn much attention \cite{Glavas2016UnsupervisedTS,Lukasik2020TextSB}.

\paragraph{\textbf{Optimal Transport}}
Optimal Transport (OT) is a field of mathematics that studies the geometry of probability spaces \cite{Villani2003TopicsIO}, which is a formalism for finding and quantifying the movement of mass from one probability distribution to another \cite{Zhu2021FunctionalOT}. The theoretical importance of OT is that it defines the Wasserstein metric between probability distributions. It reveals a canonical geometric structure with rich properties to be exploited. The earliest contribution to OT originated from Monge in the eighteenth century. Kantorovich rediscovered it under a different formalism, namely the Linear Programming formulation of OT. With the development of scalable solvers, OT is widely applied to many real-world problems \cite{Zhu2021FunctionalOT,Qiu2022OptimalTB,Flamary2021POTPO,Chen2020GraphOT,Yuan2020AdvancingWS,Klicpera2021ScalableOT,Alqahtani2021UsingOT,Lee2019HierarchicalOT}.

\paragraph{\textbf{Multimodal Alignment}}
Aligning representations from different modalities is an important
step in multimodal learning. 
With the recent advancement in computer vision and natural language processing, multimodal learning, which aims to explore the explicit relationship across vision and language, has drawn significant attention \cite{Wang2020AnEF}. 
There are many methods proposed for exploring the multimodal alignment objective.
\cite{Torabi2016LearningLE,Yu2017EndtoEndCW} adopted attention mechanisms, \cite{Dong2021DualEF} composed pairwise joint representation, \cite{Chen2020FineGrainedVR,Wray2019FineGrainedAR,Zhang2018CrossModalAH} learned fine-grained or hierarchical alignment,  \cite{Lee2018StackedCA,Wu2019UnifiedVE} decomposed the images and texts into sub-tokens, \cite{Velickovic2018GraphAN,Yao2018ExploringVR} adopted graph attention for reasoning, and \cite{Yang2021TACoTC} applied contrastive learning algorithms for video-text alignment.

\begin{figure*}[htp]
  \centering
  \includegraphics[width=1\linewidth]{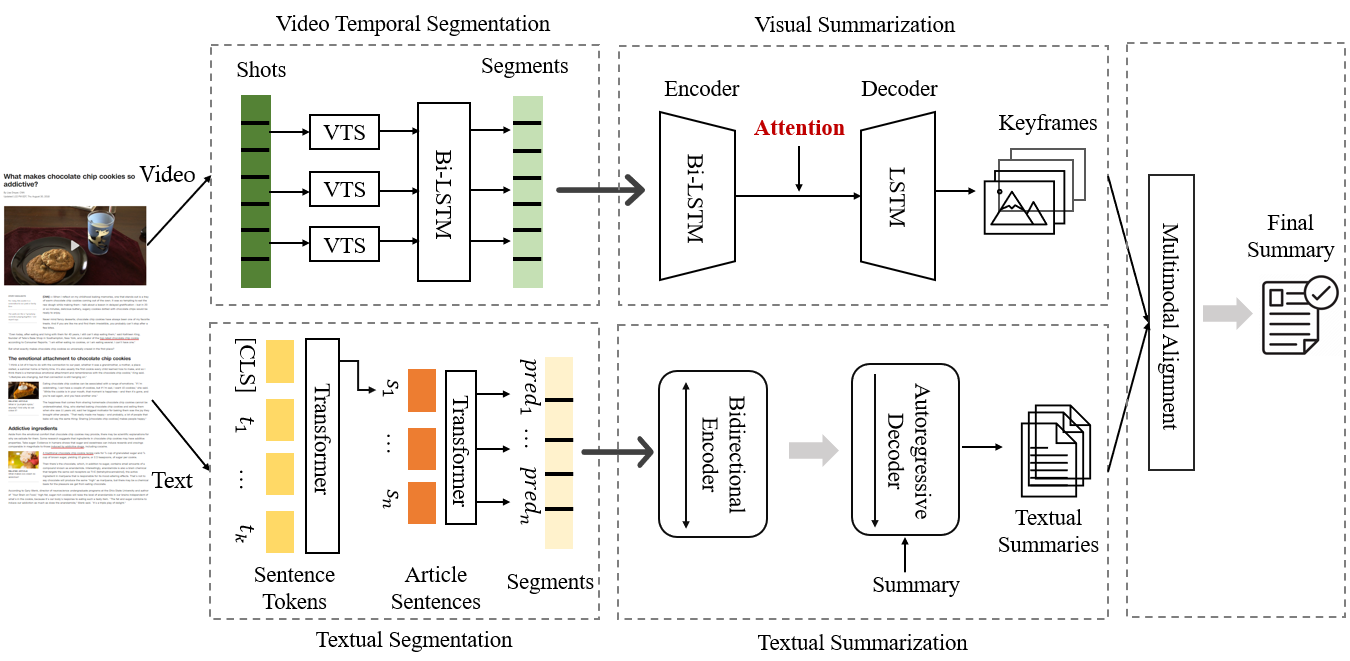}
  \caption{The framework of our MHMS model, which takes a multimedia input (video+text) and generates multimodal summaries. The framework includes five modules for: video temporal segmentation, visual summarization, textual segmentation, textual summarization, and multimodal alignment.}
  \label{fig:MHMS}
\end{figure*}

\section{Methods}
In the task of multimodal multimedia summarization with multimodal output, we need to processes visual and language information and produce both visual and language summaries. We follow the problem setting in \cite{Mingzhe2020VMSMOLT}. Given a multimedia source which contains documents/textual language and videos, the document $X_D=\{x_1, x_2, ..., x_{d} \}$ has $d$ words, and the ground truth textual summary $Y_D = \{ y_1, y_2, ..., y_{g} \}$ has $g$ words. The corresponding video $X_V$ is aligned with the document, and there exists a ground truth cover picture $Y_V$ that can represent the most important information to describe the video. For the input document and video, our model learns the joint representation of both domains to generate both textual summary $Y_D'$ and video key frame $Y_V'$, which aim at preserving the most important textual and visual information to represent the video, respectively. 

We propose a Multimodal Hierarchical Multimedia Summarization (MHMS) method, which consists of five modules: video temporal segmentation (Section~\ref{sec:video_temp}), visual summarization (Section~\ref{sec:visual_sum}), textual segmentation (Section~\ref{sec:text_seg}), textual summarization (Section~\ref{sec:text_sum}), and multimodal summarization by cross-domain alignment (Section~\ref{sec:align}).
The overall framework is shown in Figure~\ref{fig:MHMS}, where the example input is from CNN news\footnote{https://www.cnn.com/2018/08/30/health/chocolate-chip-cookies-addictive-food-drayer/index.html}. 
Each module will be introduced in the following subsections. 

\subsection{Video Temporal Segmentation}\label{sec:video_temp}
\begin{figure}[htp]
  \centering
  \includegraphics[width=1\linewidth]{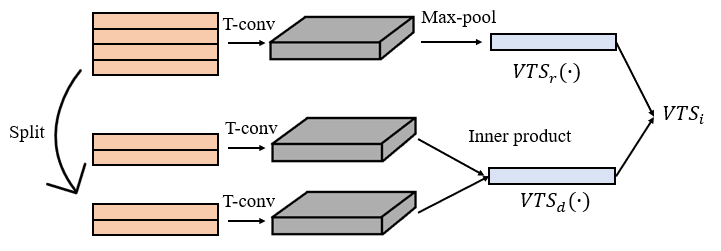}
  \caption{The VTS model in the video temporal segmentation module \cite{Rao2020ALA,PySceneDetect}.}
  \label{Fig:VTS}
\end{figure}

Video temporal segmentation (VTS) aims at splitting the original video into small segments, which the summarization tasks build upon.
Our VTS model is similar to \cite{Rao2020ALA,PySceneDetect}. 
VTS is formulated as a binary classification problem on the segment  boundaries \cite{Rao2020ALA}. Given a video $X_V$, the task for the video temporal segmentation is to separate the video sequence into scenes $[X_{v1},X_{v2}, ..., X_{vn}]$, where $n$ is the number of scenes. The VTS module gives a sequence of predictions $[P_{v1},P_{v2},...,P_{vn}]$, where $P_{vi} \in \{0, 1\}$ denotes whether the boundary between the $i$-th and $(i + 1)$-th shots is a scene boundary. The model of VTS is shown in Figure~\ref{Fig:VTS}.

We first use \cite{PySceneDetect} to split the video into shots $[S_{v1},S_{v2},...,S_{vn}]$. The model $\text VTS$ takes a clip of the video with $2 \omega_b$ shots as input and outputs a boundary representation $\text VTS_i$. The boundary representation captures both differences and relations between the shots before and after, the model $\text VTS$ consists of two branches, $\text VTS_d$ and $\text VTS_r$, which is shown in Equation~\ref{Equation:VTS_i}. $\text VTS_d$ is modeled by two temporal convolution layers, each of which embeds the $w_b$ shots before and after the boundary, respectively, following an inner product operation to calculate their differences. $\text VTS_r$ aims to capture the relations of the shots, it is implemented by a temporal convolution layer followed a max pooling. Then it predicts a sequence binary labels $[P_{v1},P_{v2},...,P_{vn}]$ based on the sequence of representatives $\text [VTS_1, VTS_2, ..., VTS_n]$. A Bi-LSTM \cite{Graves2005FramewisePC} is used with stride $\omega_t / 2$ shots to predict a sequence of coarse score $[s_1,s_2,...,s_n]$, as shown in Equation~\ref{Equation:prb}, where $s_i \in [0,1]$ is the probability of a shot boundary to be a scene boundary. The coarse prediction $\hat{P}_{vi} \in \{0, 1\}$ indicates whether the $i$-th shot boundary is a scene boundary. By binarizing $s_i$ with a threshold $\tau$, we get Equation~\ref{Equation:VTS}.

\begin{equation}
\begin{aligned}
{\text VTS}_{i} &={\text VTS}\left(\left[{S}_{vi-\left(\omega_{b}-1\right)}, \cdots, {S}_{vi+\omega_{b}}\right]\right)  \\
&=\left[\begin{array}{l}
\text{ VTS}_{d}\left(\left[{S}_{vi-\left(\omega_{b}-1\right)}, \cdots, \text{P}_{vi}\right],\left[{S}_{v(i+1)}, \cdots, {S}_{vi+\omega_{b}}\right]\right) \\
\text{ VTS}_{r}\left(\left[{S}_{vi-\left(\omega_{b}-1\right)}, \cdots, {P}_{vi}, {S}_{v(i+1)}, \cdots, {S}_{vi+\omega_{b}}\right]\right)
\end{array}\right]
\end{aligned} 
\label{Equation:VTS_i}
\end{equation}

\begin{equation}
[s_{1},s_{2},...,s_{n}]=\text{Bi-LSTM}\left(\left[{VTS}_{1}, {VTS}_{2},\cdots, {VTS}_{n}\right]\right)
\label{Equation:prb}
\end{equation}

\begin{equation}
\hat{P}_{vi}= \begin{cases}1 & \text { if } s_{i}>\tau \\ 0 & \text { otherwise }\end{cases}
\label{Equation:VTS}
\end{equation}

\subsection{Visual Summarization}\label{sec:visual_sum}
The visual summarization module extracts visual keyframes from each segment as its corresponding summary. The keyframes should be the representative frames of a video stream, which provide the most accurate and compact summary of the video content. 
We use a encoder-decoder architecture with attention as the visual summarization module \cite{Ji2020VideoSW}, which formulate video summarization as a sequence-to-sequence learning problem. The input is each video segment and the output is a sequence of keyframes.  The encoder in the visual summarization module is a Bi-LSTM \cite{Graves2005FramewisePC} to model the  temporal relationship of video frames, where the input is $X=[x_1, x_2, ..., x_m]$ and the encoding representation is $E =[e_1, e_2, ... e_m]$. The decoder is a LSTM \cite{Hochreiter1997LongSM} to learn the long term and short term dependency among the importance scores to generate the output sequence $D=[d_1, d_2, ..., d_m]$. To exploit the temporal ordering across the entire video, we introduce the  attention mechanism:
\begin{equation}
\begin{gathered}
\mathrm{E}_{t}=\sum_{i=1}^{m} \alpha_{t}^{i} e_{i}, \text { s.t. } \sum_{i=1}^{n} \alpha_{t}^{i}=1 
\end{gathered}
\end{equation}
\begin{equation}
\begin{gathered}
{\left[\begin{array}{c}
p\left(d_{t} \mid\left\{d_{i} \mid i<t\right\}, E_{t}\right) \\
s_{t}
\end{array}\right]=\psi\left(s_{t-1}, d_{t-1}, E_{t}\right)}
\end{gathered}
\end{equation}
where $s_t$ is the hidden state, $E_t$ is the attention vector at time $t$, $\alpha_t^i $ is the attention weight between the inputs and the encoder vector, $\psi$ is the decoder function. The attention weight $\alpha_t^i$ is computed at each time step $t$, which reflects the attention degree of the $i$-th temporal feature in the input video. To obtain $\alpha_t^i $, the relevance score $e_t^i$ is computed:
\begin{equation}
e_t^i =\text{score} (s_{t-1}, e_i)
\end{equation}
where the $\text{score}$ function  decides the relationship between
the $i$-th visual features $e_i$ and the output scores at time $t$, which is computed in a multiplicative way:
\begin{equation}
\beta_t^i = e^T_i W_a s_{t-1}
\end{equation}
\begin{equation}
\alpha^i_t = \exp(\beta_t^i) / \sum^m_{j=1} \exp(\beta_t^j)
\end{equation}

\subsection{Textual Segmentation}\label{sec:text_seg}
The textual segmentation module takes the whole document or articles as input and outputs the segmentation results based on the textual understanding. 
We used a hierarchical BERT as the video temporal segmentation model \cite{lukasik-etal-2020-text}. The hierarchical BERT contains two-level transformer encoders, where the first-level encoder is for sentence-level encoding, and the second-level encoder is for the article-level encoding. The hierarchical BERT starts by encoding each sentence with $\text{BERT}_{\text{LARGE}}$ independently. Then the tensors produced for each sentence are fed into another transformer encoder to capture 
the representation of the sequence of sentences. All the sequences start with a [CLS] token to encode each sentence with BERT at the first level. If the segmentation decision is made at the sentence level, we use the [CLS] token as input of the second-level encoder. The [CLS] token representations from sentences are passed into the article encoder, which can relate the different sentences through cross-attention.

Due to the quadratic computational cost of transformers, we reduce the BERT’s inputs to 64 word-pieces per sentence and 128 sentences per document like \cite{lukasik-etal-2020-text}. We use 12 layers for both the sentence and the article encoders, for a total of 24 layers. In order to use the $\text{BERT}_{\text{BASE}}$ checkpoint, we use 12 attention heads and 768-dimensional word-piece embeddings.

\subsection{Textual Summarization}\label{sec:text_sum}
Language summarization can produce a concise and fluent summary which should preserve the critical information and overall meaning. 
To generate a more accurate summary, we take the abstractive summarization method in our pipeline. 
Our textual summarization module takes Bidirectional and Auto-Regressive Transformers (BART) \cite{Lewis2020BARTDS} as the summarization model to generate abstractive textual summary candidates. BART is a denoising autoencoder that maps a corrupted document to the original document it was derived from. It is implemented as a sequence-to-sequence model with a bidirectional encoder over corrupted text and a left-to-right autoregressive decoder. BART uses the standard sequence-to-sequence Transformer architecture, where both the encoder and the decoder include 12 layers. In addition to the stacking of encoders and decoders, cross attention between encoder and decoder is also applied. BART is trained by corrupting documents and then optimizing a reconstruction loss, where the pretraining task involves randomly shuffling the order of the original sentences and a novel in-filling scheme, where spans of text are replaced with a single mask token. BART is particularly effective when fine tuned for text generation but also works well for comprehension tasks, including achieving new state-of-the-art results on the summarization task.

\subsection{Cross-Domain Alignment for Multimodal Summarization}\label{sec:align}
The final module is to learn the relationship and alignment between keyfames and textual summaries to generate the final results. Our alignment module is based on Optimal Transport (OT), which has been explored in several cross-domain tasks 
\cite{Chen2020GraphOT,Yuan2020AdvancingWS,Lu2021CrossdomainAR}.

\begin{figure}[htp]
  \centering
  \includegraphics[width=0.8\linewidth]{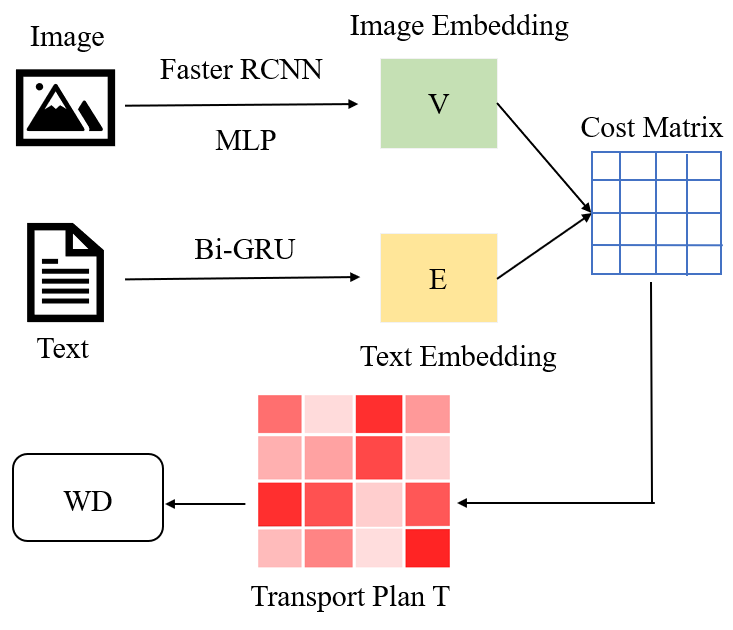}
  \caption{The multimodal alignment module.}
  \label{Fig:WD}
\end{figure}

Our multimodal alignment module is shown in Figure~\ref{Fig:WD}, which is inspired by OT \cite{Yuan2020AdvancingWS}.  OT is the problem of transporting mass between two discrete distributions supported on some latent feature space $\mathcal{X}$. Let $\boldsymbol{\mu}=\left\{\boldsymbol{x}_{i}, \boldsymbol{\mu}_{i}\right\}_{i=1}^{n}$ and $\boldsymbol{v}=\left\{\boldsymbol{y}_{j}, \boldsymbol{v}_{j}\right\}_{j=1}^{m}$ be the discrete distributions of interest, where $\boldsymbol{x}_{i}, \boldsymbol{y}_{j} \in \mathcal{X}$ denotes the spatial locations and $\mu_{i}, v_{j}$, respectively, denoting the non-negative masses. Without loss of generality, we assume $\sum_{i} \mu_{i}=\sum_{j} v_{j}=1$. $\pi \in \mathbb{R}_{+}^{n \times m}$ is a valid transport plan if its row and column marginals match $\mu$ and $\boldsymbol{v}$, respectively, which is $\sum_{i} \pi_{i j}=v_{j}$ and $\sum_{j} \pi_{i j}=\mu_{i}$. Intuitively, $\pi$ transports $\pi_{i j}$ units of mass at location $\boldsymbol{x}_{i}$ to new location $\boldsymbol{y}_{j}$. Such transport plans are not unique, and one often seeks a solution $\pi^{*} \in \Pi(\boldsymbol{\mu}, \boldsymbol{v})$ that is most preferable in other ways, where $\Pi(\boldsymbol{\mu}, \boldsymbol{v})$ denotes the set of all viable transport plans. OT finds a solution that is most cost effective w.r.t. some function $C(\boldsymbol{x}, \boldsymbol{y})$:
\begin{equation}
\mathcal{D}(\boldsymbol{\mu}, \boldsymbol{v})=\sum_{i j} \pi_{i j}^{*} C\left(\boldsymbol{x}_{i}, \boldsymbol{y}_{j}\right)=\inf _{\pi \in \Pi(\mu, v)} \sum_{i j} \pi_{i j} C\left(\boldsymbol{x}_{i}, \boldsymbol{y}_{j}\right)
\end{equation}
where $\mathcal{D}(\boldsymbol{\mu}, \boldsymbol{v})$ is known as the optimal transport distance. Hence, $\mathcal{D}(\boldsymbol{\mu}, \boldsymbol{v})$ minimizes the transport cost from $\boldsymbol{\mu}$ to $\boldsymbol{v}$ w.r.t. $C(\boldsymbol{x}, \boldsymbol{y})$. When $C(\boldsymbol{x}, \boldsymbol{y})$ defines a distance metric on $\mathcal{X}$, and $\mathcal{D}(\boldsymbol{\mu}, \boldsymbol{v})$ induces a distance metric on the space of probability distributions supported on $\mathcal{X}$, it becomes the Wasserstein Distance (WD).

The image features $V=\left\{\boldsymbol{v}_{k}\right\}_{k=1}^{K}$ are extracted from a pre-trained ResNet-101 \cite{He2016DeepRL} concatenated to faster R-CNN \cite{Ren2015FasterRT} as \cite{Yuan2020AdvancingWS}. For text features, every word (token) is first embedded as a feature vector, and processed by a bi-directional Gated Recurrent Unit (Bi-GRU) \cite{Schuster1997BidirectionalRN} to account for context \cite{Yuan2020AdvancingWS}. The extracted image and text embeddings are $\mathbf{E}=\left\{\boldsymbol{e}_{i}\right\}_{1}^{M}, \mathbf{V}=\left\{\boldsymbol{v}_{i}\right\}_{1}^{K}$, respectively.

We take image and text sequence embeddings as two discrete distributions supported on the same feature representation space. Solving an OT transport plan between the two naturally constitutes a matching scheme to relate cross-domain entities \cite{Yuan2020AdvancingWS}. To evaluate the OT distance, we compute a pairwise similarity between $V$ and $E$
using cosine distance:
\begin{equation}
\begin{aligned}
C_{km} = C(e_k, v_m) = 1-\frac{\boldsymbol{e}_{k}^{T} \boldsymbol{v}_{k}}{\left\|\boldsymbol{e}_{k}\right\|\left\|\boldsymbol{v}_{m}\right\|}
\end{aligned}
\end{equation}
Then the OT can be formulated as:
\begin{equation}
\label{eq:ot_lp}
\mathcal{L}_{\mathrm{OT}}(\mathbf{V}, \mathbf{E})=\min _{\mathbf{T}} \sum_{k=1}^{K} \sum_{m=1}^{M} \mathbf{T}_{k m} \mathbf{C}_{k m}
\end{equation}
where $\sum_{m} \mathbf{T}_{k m}=\mu_{k}, \sum_{k} \mathbf{T}_{k m}=v_{m}, \forall k \in[1, K], m \in[1, M] .$ and $\mathbf{T} \in \mathbb{R}_{+}^{K \times M}$ is the transport matrix, $d_{k}$ and $d_{m}$ are the weight of $\boldsymbol{v}_{k}$ and $\boldsymbol{e}_{m}$ in a given image and text sequence, respectively. We assume the weight for different features to be uniform, i.e., $\mu_{k}=\frac{1}{K}, v_{m}=\frac{1}{M}$. 
The objective of optimal transport involves solving linear programming and may cause potential computational burdens since it has $O(n^3)$ efficiency. To solve this issue, we add an entropic regularization term equation (\ref{eq:ot_lp}) and the objective of our optimal transport distance becomes
\begin{align}
    \mathcal{L}_{\mathrm{OT}}(\mathbf{V}, \mathbf{E})=\min _{\mathbf{T}} \sum_{k=1}^{K} \sum_{m=1}^{M} \mathbf{T}_{k m} \mathbf{C}_{k m} + \lambda H(\mathbf{T}),
\end{align}
where $H(\mathbf{T}) = \sum_{i,j} \mathbf{T}_{i,j} \log \mathbf{T}_{i,j}$ is the entropy, and $\lambda$ is the hyperparameter that balance the effect of the entropy term. Thus, we are able to apply the celebrated Sinkhorn algorithm \cite{cuturi2013sinkhorn} to efficiently solve the above equation in $O(n 
log n)$, where the algorithm is shown in Algorithm~\ref{alg:WD}.  The optimal transport distance computed via the Sinkhorn algorithm is differentiable and it can be implemented with deep learning libraries \cite{Flamary2021POTPO}.
After we train the alignment module,  we are able to compute the WD between each keyframe-sentence pair of all the visual \& textual summary candidates, which enable us to select the best match as the final multimodal summaries.

\begin{algorithm}[htp] 
	\caption{Compute Multimodal Alignment Distance}
	\begin{algorithmic}[1]
		\State \textbf{Input}: $ \mathbf{V}=\left\{\boldsymbol{v}_{i}\right\}_{1}^{K}, 
		\mathbf{E}=\left\{\boldsymbol{e}_{i}\right\}_{1}^{M},\beta$
		\State $\mathbf{C}=C(\mathbf{V}, \mathbf{E})$, $\sigma \leftarrow \frac{1}{m} \mathbf{1}_{m}, \mathbf{T}^{(1)} \leftarrow \mathbf{1} \mathbf{1}^{T}$
		\State $\mathbf{G}_{i j} \leftarrow \exp \left(-\frac{\boldsymbol{C}_{i j}}{\beta}\right)$
		\For{ t = 1,2,3,...,N}
		\State$\boldsymbol{Q} \leftarrow \boldsymbol{G} \odot \mathbf{T}^{(t)}$
		\For{l = 1,2,3,...,L}
		\State $\boldsymbol{\delta} \leftarrow \frac{1}{K \boldsymbol{Q} \sigma}, \boldsymbol{\sigma} \leftarrow \frac{1}{M \boldsymbol{Q}^{T} \boldsymbol{\delta}}$
		\EndFor
        \State $\mathbf{T}^{(t+1)} \leftarrow \operatorname{diag}(\boldsymbol{\delta}) \boldsymbol{Q} \operatorname{diag}(\boldsymbol{\sigma})$
        \EndFor
        \State $\mathbf{Dis} = <C^T, T>$
	\end{algorithmic}
	\label{alg:WD}
\end{algorithm}

\section{Datasets and Baselines}
\subsection{Datasets}\label{sec:dataset}
We evaluated our models on three datasets: VMSMO dataset \cite{Mingzhe2020VMSMOLT},  Daily Mail dataset, and CNN dataset from \cite{Mingzhe2020VMSMOLT,Fu2021MMAVSAF,Fu2020MultimodalSF}. The popular COIN and Howto100M can not be used in our task, since they lack narrations and key-step annotation \cite{Tang2019COINAL,miech19howto100m}.

The VMSMO dataset contains  184,920 samples, including articles and corresponding videos. Each sample is assigned with a textual summary and a video with a cover picture. We adopted the available data samples from \cite{Mingzhe2020VMSMOLT} and replaced the unavailable ones following the same procedure as \cite{Mingzhe2020VMSMOLT}. The Daily Mail dataset contains 1,970 samples, and the CNN dataset contains 203 samples, where they both include video titles, images, and their captions, which are similar to \cite{Hermann2015TeachingMT}.

For the data splitting, we take the same experimental setup as \cite{Mingzhe2020VMSMOLT} for the VMSMO dataset. For Daily Mail dataset and CNN dataset, we split the data by 70\%, 10\%, 20\% for train, validation and test sets, respectively, same as \cite{Fu2021MMAVSAF,Fu2020MultimodalSF}.

\subsection{Baselines}
\subsubsection{Baselines for the VMSMO dataset}
For the VMSMO dataset, we compare with multimodal summarization baselines and textual summarization baselines:

\noindent \underline{\em Multimodal summarization baselines:}\\
\textbf{Synergistic} \cite{Guo2019ImageQuestionAnswerSN}: \cite{Guo2019ImageQuestionAnswerSN} proposed a image-question-answer synergistic network to value the role of the answer for precise visual dialog, which is able to jointly learn the representation of the image, question, answer, and history in a single step. \\
\textbf{PSAC} \cite{Li2019BeyondRP}: The Positional Self-Attention with Coattention (PSAC) model adopted positional self-attention block to model the data dependencies and video-question co-attention to help attend to both visual and textual information. \\
\textbf{MSMO} \cite{Zhu2018MSMOMS}: MSMO was the first model on producing  multimodal output as summarization results, which adopted the pointer-generator network, added attention to text and images when generating textual summary, and used visual coverage by the sum of visual attention distributions to select pictures.\\
\textbf{MOF} \cite{Zhu2020MultimodalSW}: \cite{Zhu2020MultimodalSW} proposed a multimodal objective function with the guidance of multimodal reference
to use the loss from the summary generation and the image selection to solve the modality-bias problem.\\
\textbf{DIMS} \cite{Mingzhe2020VMSMOLT}: DIMS is a dual interaction module and multimodal generator, where conditional self-attention mechanism is used to capture local semantic information within video, and the global-attention mechanism is applied to handle the semantic relationship between news text and video from a high level.

\noindent \underline{\em Textual summarization baselines:}\\
\textbf{Lead} \cite{Nallapati2017SummaRuNNerAR}:  The Lead method simply selects the first sentence of article/document as the textual summary.\\
\textbf{TexkRank} \cite{Mihalcea2004TextRankBO}: TexkRank is a graph-based extractive summarization method which adds sentences as nodes and uses edges to weight similarity.\\
\textbf{PG} \cite{Abigail2017}: PG is a hybrid  pointer-generator model  with coverage, which  copied words via pointing, and generated words from a fixed vocabulary with attention. \\
\textbf{Unified} \cite{Hsu2018AUM}: The Unified model combined the strength of extractive and abstractive summarization, where a sentence-level attention is used to modulate the word-level attention and an inconsistency loss function is introduced to penalize the inconsistency between two levels of attentions.\\
\textbf{GPG} \cite{Shen2019ImprovingLA}: Generalized Pointer Generator (GPG) replaced the hard copy component with a more general soft “editing” function, which  learns a relation embedding to transform the pointed word into a target embedding.

\subsubsection{Baselines for Daily Mail  and CNN datasets} For Daily Mail  and CNN datasets, we have multimodal baselines, video summarization baselines, and textual summarization baselines:

\noindent \underline{\em Multimodal summarization baselines:}\\
\textbf{VistaNet} \cite{vistanet}: VistaNet relies on visual information as alignment for pointing out the important sentences of a document using attention to detect the sentiment expressed by a document.\\
\textbf{MM-ATG} \cite{Zhu2018MSMOMS}: MM-ATG is a multi-modal attention on global features (ATG) model to generate text and select the relevant image from the article and alternative images.\\
\textbf{Img+Trans} \cite{hori2019end}: \cite{hori2019end} applied multi-modal video features including video frames, transcripts, and dialog context for dialog generation.\\
\textbf{TFN} \cite{zadeh-etal-2017-tensor}: Tensor Fusion Network (TFN) models intra-modality and inter-modality dynamics for f multimodal sentiment analysis which explicitly represents unimodal, bimodal, and trimodal interactions between behaviors.\\
\textbf{HNNattTI} \cite{chen2018abstractive}: HNNattTI aligned the sentences and accompanying images by using attention mechanism. \\
\textbf{M$^2$SM} \cite{Fu2021MMAVSAF,Fu2020MultimodalSF}: M$^2$SM is a multimodal summarization model with a bi-stream summarization strategy for training by sharing the
ability to refine significant information from long materials in text and video summarization.

\noindent \underline{\em Video summarization baselines:}\\
\textbf{VSUMM} \cite{de2011vsumm}: VSUMM is a methodology for the production of static video summaries, which extracted color features from video frames and adopted k-means for clustering.\\
\textbf{Random}: The Random method means extracting the key video frames randomly as the summarization result.\\
\textbf{Uniform}: The Uniform method means sampling the videos uniformly for keyframe selection as video summaries. \\
\textbf{DR-DSN} \cite{zhou2018deep}: \cite{zhou2018deep} formulated video summarization as a sequential decision making process and developed a deep summarization network (DSN) to summarize videos. DSN predicted a probability for each frame, which indicates the likelihood of a frame being selected, and then takes actions based on the probability distributions to select frames to from video summaries.

\noindent \underline{Textual summarization baselines:}\\
\textbf{Lead3}: Similar to Lead, Lead3 means picking the first three sentences as the summary result.\\
\textbf{SummaRuNNer} \cite{Nallapati2017SummaRuNNerAR}: SummaRuNNer is a RNN-based sequence model for extractive summarization of documents, which used abstractive training on human generated reference summaries to eliminate the need for sentence-level extractive labels.\\
\textbf{NN-SE} \cite{cheng2016neural}: NN-SE is a general framework for single-document summarization composed of a hierarchical document encoder and an attention-based extractor.

\section{Experiments}
\subsection{Implementation}
\paragraph{Video Temporal Segmentation} We used the same model setting as \cite{Rao2020ALA,PySceneDetect} and same data splitting setting  as \cite{Mingzhe2020VMSMOLT,Fu2021MMAVSAF,Fu2020MultimodalSF} to train the video temporal segmentation module.

\paragraph{Visual Summarization} The visual summarization model is pre-trained on the TVSum \cite{Song2015TVSumSW} and SumMe \cite{Gygli2014CreatingSF} datasets. TVSum dataset contains 50 edited videos downloaded from YouTube in 10 categories, and SumMe dataset consists of 25 raw videos recording various events, where frame-level importance scores for each video are provided for both datasets, which are used as ground-truth labels. The input visual features are extracted from pre-trained GoogLeNet on ImageNet, where the output of the pool5 layer is used as visual features.

\paragraph{Textual Segmentation} The hierarchical BERT model is pre-trained on the Wiki-727K dataset \cite{Koshorek2018TextSA}, which contains 727 thousands articles from a snapshot of the English Wikipedia. We used the same data splitting method as \cite{Koshorek2018TextSA}.  

\paragraph{Textual Summarization} We used the BART \cite{Lewis2020BARTDS} as the abstractive textual summarization model. We adopted the pretrained BART model (bart-large-cnn\footnote{https://huggingface.co/facebook/bart-large-cnn}) from \cite{Lewis2020BARTDS}, which contains 1024 hidden layers and 406M parameters and has been fine-tuned using CNN and Daily Mail datasets. 

\paragraph{Multimodal Alignment} The feature extraction and alignment module is pretrained by MS COCO dataset \cite{Lin2014MicrosoftCC} on the image-text matching task. We added the OT loss as a regularization term to the original matching loss to align the image and text more explicitly.

\subsection{Experiments and Results}
The quality of generated textual summary is evaluated by standard full-length Rouge F1 \cite{Lin2004ROUGEAP} following previous works \cite{Abigail2017,Chen2018IterativeDR,Mingzhe2020VMSMOLT}.  ROUGE-1 (R-1), ROUGE-2 (R-2), and ROUGE-L (R-L) refer to overlap of unigram, bigrams, and the longest common subsequence between the decoded summary and the reference, respectively \cite{Lin2004ROUGEAP}. 

For VMSMO dataset, the quality of chosen cover frame is evaluated by mean average precision (MAP) and recall at position $(R_n @ k)$ \cite{Zhou2018MultiTurnRS,Tao2019MultiRepresentationFN}, where $(R_n @ k)$ measures if the positive sample is ranked in the top $k$ positions of $n$ candidates. For Daily Mail  dataset and CNN dataset, we calculate the cosine image similarity (Cos) between image references and
the extracted frames from videos \cite{Fu2021MMAVSAF,Fu2020MultimodalSF}.

\begin{table}[htp]\small
    \centering
	\caption{Comparison with multimodal baselines on the VMSMO dataset.}
	\begin{tabular}{lccccccc}  
			\hline
			 \multirow{2}*{Methods}   & \multicolumn{3}{c}{Textual} &\multicolumn{4}{c}{Video}   \\ 
			    \cmidrule(r){2-4}\cmidrule(r){5-8}
			         &R-1 &R-2 &R-L & MAP &$R_{10}@1$ &$R_{10}@2$ &$R_{10}@5$ \\  \hline
			MSMO \cite{Zhu2018MSMOMS} & 20.1 &4.6 &17.3 &0.554 &0.361 &0.551 &0.820   \\
			MOF \cite{Zhu2020MultimodalSW}  &21.3 &5.7 &17.9 &0.615 &0.455 &0.615 &0.817    \\
			DIMS \cite{Mingzhe2020VMSMOLT} &25.1 &9.6 &23.2 &0.654 &0.524 &0.634 &0.824 \\
			Ours     &\textbf{27.1}   &\textbf{9.8}  &\textbf{25.4}    &\textbf{0.693}   &\textbf{0.582}   &\textbf{0.688}   &\textbf{0.895}   \\ \hline
	\end{tabular}
	\label{table:VMSMO_all}
\end{table}

\begin{table}[htp]
    \centering
	\caption{Comparison of video summarization baselines on the VMSMO dataset.}
		\begin{tabular}{lcccc}  
			\hline
			 Method        & MAP &$R_{10}@1$ &$R_{10}@2$ &$R_{10}@5$   \\  \hline
			Synergistic \cite{Guo2019ImageQuestionAnswerSN}& 0.558 &0.444 & 0.557 & 0.759 \\
			PSAC \cite{Li2019BeyondRP}  &0.524 &0.363 &0.481 &0.730    \\ 
			Ours     &\textbf{0.693}   &\textbf{0.582}   &\textbf{0.688}   &\textbf{0.895}     \\ \hline
	\end{tabular}
	\label{table:VMSMO_visual}
\end{table}

\begin{table}[htp]
    \centering
	\caption{Comparison with traditional textual summarization baselines on the VMSMO dataset.}
		\begin{tabular}{lccc}  
			\hline
			Method         &R-1 &R-2 &  R-L    \\  \hline
	      	Lead \cite{Nallapati2017SummaRuNNerAR}    & 16.2  &5.3  &13.9     \\ 
			TextRank \cite{Mihalcea2004TextRankBO} & 13.7  & 4.0 & 12.5    \\   
			PG  \cite{Abigail2017}     & 19.4  & 6.8 & 17.4    \\
			Unified \cite{Hsu2018AUM} & 23.0  & 6.0 & 20.9    \\
			GPG   \cite{Shen2019ImprovingLA}   & 20.1  & 4.5 & 17.3    \\
			DIMS \cite{Mingzhe2020VMSMOLT}  & 25.1  & 9.6  &  23.2 \\
			Ours    &\textbf{27.1}   &\textbf{9.8}  &\textbf{25.4}     \\ \hline
	\end{tabular}
	\label{table:VMSMO_text}
\end{table}

We compare our MHMS model with existing multimodal summarization, video summarization, and textual summarization approaches. The comparison results on the VMSMO dataset of multimodal, video, and textual summarization are shown in Table~\ref{table:VMSMO_all}, Table~\ref{table:VMSMO_visual}, Table~\ref{table:VMSMO_text}, respectively. 
Synergistic \cite{Guo2019ImageQuestionAnswerSN} and  PSAC \cite{Li2019BeyondRP} are video pure summarization approaches, which did not perform as good as multimodal methods, like MOF \cite{Zhu2020MultimodalSW} or DIMS \cite{Mingzhe2020VMSMOLT}, which means taking the additional modality into consideration actually helps to improve the quality of the generated video summaries. 
Our MHMS method is able to align more matched keyframes with textual deceptions, which shows better performance than the previous ones.
If comparing the quality of generated textual summaries, our method still outperforms the other multimodal baselines, like MSMO \cite{Zhu2018MSMOMS}, MOF \cite{Zhu2020MultimodalSW}, DIMS \cite{Mingzhe2020VMSMOLT}, and also trationanl textual summarization methods, like Lead \cite{Nallapati2017SummaRuNNerAR}, TextRank \cite{Mihalcea2004TextRankBO}, PG  \cite{Abigail2017}, Unified \cite{Hsu2018AUM},  and GPG   \cite{Shen2019ImprovingLA}, 
showing the alignment  obtained by optimal transport can help to identify the cross-domain inter-relationships.

In Table~\ref{table:Dailymail_CNN_all}, we show the comparison results with multimodal baselines on the Daily Mail  and CNN datasets. 
We can see that for the CNN datasets, our method shows competitive results with
Img+Trans \cite{hori2019end}, TFN \cite{zadeh-etal-2017-tensor}, HNNattTI \cite{chen2018abstractive} and $\rm M^{2}$SM \cite{Fu2021MMAVSAF} on the quality of generated textual summaries. 
While on the Daily Mail dataset, our MHMS approach showed better performance on both textual summaries and visual summaries. 
We also compare with the traditional pure video summarization baselines \cite{de2011vsumm,zhou2018deep,Fu2021MMAVSAF} and pure textual summarization baselines \cite{Nallapati2017SummaRuNNerAR,cheng2016neural} on the Daily Mail dataset, and the results are shown in Table~\ref{table:dailymail_visual} and Table~\ref{table:dailymail_text}, respectively. We can find that the quality of generated visual summary by our approach still outperforms the other visual summarization baselines. As for textual summarization comparison, our approach performed competitive results compared with NN-SE \citep{cheng2016neural}  and $\rm M^{2}$SM \cite{Fu2021MMAVSAF}.

\begin{table}[htp]\small
\centering
\caption{Comparisons of multimodal baselines on the Daily Mail and CNN datasets.}
\begin{tabular}{lccccccc} \hline
\multirow{2}*{Methods} & \multicolumn{3}{c}{CNN dataset} & \multicolumn{4}{c}{Daily Mail dataset}   \\ \cmidrule(r){2-4}\cmidrule(r){5-8}
&R-1 &R-2 &R-L &R-1  &R-2  &R-L &Cos(\%)   \\ \hline
VistaNet \cite{vistanet} &9.31 &3.24 &6.33 &18.62 &6.77 &13.65 &-    \\ 
MM-ATG  \cite{Zhu2018MSMOMS} &26.83 &8.11 &18.34 &35.38  &14.79  &25.41 &69.17    \\
Img+Trans \cite{hori2019end}   &27.04 &8.29 &18.54 &39.28  &16.64  &28.53  &-   \\
TFN \cite{zadeh-etal-2017-tensor}  &27.68 &8.69 &18.71 &39.37  &16.38  &28.09 &-       \\  
HNNattTI \cite{chen2018abstractive} &27.61 &8.74 &18.64 &39.58  &16.71 &29.04 &68.76    \\ 
$\rm M^{2}$SM \cite{Fu2021MMAVSAF} &27.81  &8.87  &18.73 &41.73 &18.59 &31.68 &69.22     \\ 
Ours   & \textbf{28.02}   & \textbf{8.94}    & \textbf{18.89}      & \textbf{42.34}  & \textbf{19.12}  & \textbf{32.35}    & \textbf{72.45}       \\ \hline
\end{tabular}
\label{table:Dailymail_CNN_all}
\end{table}

\begin{table}[htp]
\centering
\caption{Comparison with video summarization baselines on the Daily Mail dataset. }
\begin{tabular}{lc} \hline
Model &Cos(\%)  \\ \hline
VSUMM \cite{de2011vsumm} &68.74  \\
Random &67.69   \\ 
Uniform &68.79   \\
DR-DSN \citep{zhou2018deep} &68.69   \\ 
$\rm M^{2}$SM \cite{Fu2021MMAVSAF}  &69.22  \\ 
Ours &   \textbf{72.45}               \\\hline
\end{tabular}
\label{table:dailymail_visual}
\end{table}

\begin{table}[htp]
\centering
\caption{Comparison with textual summarization baselines on the Daily Mail dataset. }
\begin{tabular}{lccc} \hline
Model &R-1  &R-2  &R-L \\ \hline
Lead3  &41.07 &17.87 &30.90 \\
SummaRuNNer \citep{Nallapati2017SummaRuNNerAR} &41.12 &17.92 &30.94 \\
NN-SE \citep{cheng2016neural}  &41.22 &18.15 &31.22 \\
$\rm M^{2}$SM \cite{Fu2021MMAVSAF} &41.73 &18.59 &31.68  \\ 
Ours & \textbf{42.34}  & \textbf{19.12}  & \textbf{32.35}  \\\hline
\end{tabular}
\label{table:dailymail_text}
\end{table}

\subsection{Ablation Study}
To evaluate each component's performance, we performed ablation experiments on different modalities and different datasets. For the VMSMO dataset, we compare the performance of using only visual information, only textual information, and multimodal information. The comparison result is shown in Table~\ref{table:VMSMO_ablation}. We also carried out experiments on different modalities using Daily Mail  dataset to show the performance of unimodal and multimodal components, and the results are shown in Table~\ref{table:dailymail_ablation}.

For the ablation experiments, when only textual data is available, we adopt BERT~\cite{Devlin2019BERTPO} to generate text embeddings and K-Means clustering to identify sentences closest to the centroid for textual summary selection. While if only video data is available, we solve the visual summarization task in an unsupervised manner, where we use K-Means clustering to cluster frames using image histogram and then select the best frame from clusters based on variance of laplacian as the visual summary.

From Table~\ref{table:VMSMO_ablation} and Table~\ref{table:dailymail_ablation}, we can find that multimodal method outperform unimodal approaches, showing the effectiveness of exploring  the relationship and taking advantage of the cross-domain alignments of generating high-quality summaries. 

\begin{table}[htp]\small
    \centering
	\caption{Ablation study to evaluate the effects of different
components on VMSMO dataset.}
	\begin{tabular}{lccccccc}  
			\hline
			    & \multicolumn{3}{c}{Textual} &\multicolumn{4}{c}{Video}   \\ 
			         &R-1 &R-2 &R-L & MAP &$R_{10}@1$ &$R_{10}@2$ &$R_{10}@5$ \\  \hline
			Ours-textual     &26.2   &9.6  &24.1  &--  &-- &-- &--   \\ 
			Ours-video      &--  & -- & --  &0.678  &0.561  &0.642  &0.863  \\ 
			Ours    &\textbf{27.1}   &\textbf{9.8}  &\textbf{25.4}     &\textbf{0.693}   &\textbf{0.582}   &\textbf{0.688}   &\textbf{0.895}   \\   \hline
	\end{tabular}
	\label{table:VMSMO_ablation}
\end{table}

\begin{table}[htp]
\centering 
\caption{Ablation study to evaluate the effects of different components on Daily Mail dataset.}
\begin{tabular}{lcccc} \hline
 &R-1 &R-2 &R-L &Cos(\%)\\ \hline
Ours-textual &40.28 &17.93 &31.89  &--\\
Ours-video &-- &-- &--  &70.56   \\  
Ours & \textbf{42.34}  & \textbf{19.12}  & \textbf{32.35}    & \textbf{72.45} \\  \hline
\end{tabular}
\label{table:dailymail_ablation}
\end{table}

\subsection{Interpretation}
\begin{figure}[htp]
  \centering
  \includegraphics[width=1\linewidth]{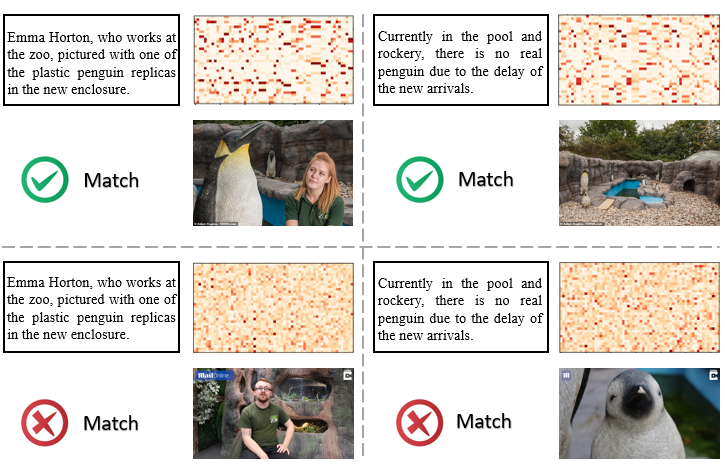}
  \caption{An illustration of the learned transport plan between visual and textual domains.}
  \label{Fig:plan}
\end{figure}

To have a deeper understanding of the multimodal alignment between the visual domain and language domain, we compute and visualize the transport plan to provide an interpretation of the latent representations, which is shown in Figure~\ref{Fig:plan}. 
When we are regarding the extracted embedding from both text and image spaces as the distribution over their corresponding spaces, we expect the optimal transport coupling to reveal the underlying similarity and structure. Also, the coupling seeks sparsity, which further helps to explain the correspondence between the text and image data.

Figure~\ref{Fig:plan} shows comparison results of matched image-text pairs and non-matched ones. The top two pairs are shown as matched pairs, where there is overlapping  between the image and the corresponding sentence. The bottom two pairs are shown as non-matched ones, where the overlapping of meaning between the image and text is relatively small. The correlation between the image domain and the language domain can be easily interpreted by the learned transport plan matrix. 
In specific, the optimal transport coupling shows the pattern of sequentially structured knowledge. However, for non-matched image-sentences pairs, the estimated couplings are relatively dense and barely contain any informative structure.

As shown in Figure~\ref{Fig:plan}, we can find that the transport plan learned in the multimodal alignment module demonstrates a way to align the features from different modalities to represent the key components. The visualization of the transport plan contributes to the interpretability of the proposed model, which brings a clear understanding of the alignment module.

\section{Conclusion and Future Work}
In this work, we proposed MHMS, a multimodal hierarchical multimedia summarization framework for generating multimodal output as summaries given multimedia sources. Our MHMS compartmentalized the algorithm into different functional modules for video temporal segmentation, textual segmentation, visual summarization, textual summarization, and multimodal alignment. The experimental results on three datasets show that MHMS outperforms previous summarization methods. Our approach provides a new direction for generating multimedia summaries, which can be extended to many real-world multimedia applications.

For future work, we are trying to expand the current work to more extended multimedia and generate multiple multimodal summaries for each section of the long multimedia. This direction will significantly improve the user experience when exploring a considerable amount of online multimedia. However, this future approach requires human annotations for organizing an extended multimedia dataset, which will be time-consuming and labor-intensive. Nevertheless, we believe the multimodal summarization task is promising and can provide valuable solutions to many real-world problems.


\clearpage
\bibliographystyle{ACM-Reference-Format}
\bibliography{reference}










\end{document}